\documentclass[letterpaper, 10 pt, conference]{ieeeconf}  

\IEEEoverridecommandlockouts                              

\usepackage{cite}
\usepackage{amsmath,amssymb,amsfonts}
\usepackage{graphicx}
\usepackage{textcomp}
\usepackage{subfigure}
\usepackage{xcolor}
\usepackage{multirow}
\usepackage{colortbl} 
\usepackage{fancyhdr}  

\title{\LARGE \bf
Crossing the Sim2Real Gap Between Simulation and Ground Testing to Space Deployment of Autonomous Free-flyer Control
}

\author{Kenneth Stewart$^{1*}$, Samantha Chapin$^{1*}$, Roxana Leontie$^{1}$, and Carl Glen Henshaw$^{1}$
\thanks{*These authors contributed equally to this work}
\thanks{$^{1}$ Authors are with the U.S. Naval Research Laboratory Naval Center for Space Technology, Washington, DC, United States
        {\tt\small kenneth.m.stewart45.civ@us.navy.mil}}
}

\begin{document}


\maketitle
\thispagestyle{empty}
\pagestyle{empty}

\begin{abstract}
Reinforcement learning (RL) offers transformative potential for robotic control in space.  
We present the first on-orbit demonstration of RL-based autonomous control of a free-flying robot, the NASA Astrobee, aboard the International Space Station (ISS). 
Using NVIDIA's Omniverse physics simulator and curriculum learning, we trained a deep neural network to replace Astrobee's standard attitude and translation control, enabling it to navigate in microgravity. 
Our results validate a novel training pipeline that bridges the simulation-to-reality (Sim2Real) gap, utilizing a GPU-accelerated, scientific-grade simulation environment for efficient Monte Carlo RL training. 
This successful deployment demonstrates the feasibility of training RL policies terrestrially and transferring them to space-based applications. 
This paves the way for future work in In-Space Servicing, Assembly, and Manufacturing (ISAM), enabling rapid on-orbit adaptation to dynamic mission requirements.
\end{abstract}

\section{Introduction} 
Future In-Space Servicing, Assembly, and Manufacturing (ISAM) missions require increasingly autonomous robotic systems capable of adapting to the dynamic and uncertain conditions of space. While traditional spacecraft control methods are often limited in their adaptability, RL offers the potential to learn robust control policies directly from experience. However, a major obstacle to deploying RL in space is the Sim2Real gap, representing the discrepancy between simulated training environments and the complexities of the real world.  This gap can lead to significant performance degradation when transferring RL policies from simulation to hardware. In this paper, we address both the need for increased autonomy in ISAM and the challenge of Sim2Real transfer by presenting the first on-orbit demonstration of RL-based control for a free-flying robot. We trained a deep neural network using NVIDIA's Omniverse physics simulator and curriculum learning to control the NASA Astrobee aboard the International Space Station (ISS). This successful deployment validates our novel training pipeline for bridging the Sim2Real gap and enabling rapid adaptation to changing mission needs, representing a significant step toward realizing the potential of autonomous robotic Assembly, Integration and Testing (AI\&T) for future space missions. We validate a deep RL (DRL) control law for the NASA Astrobee robot which we have verified via ground-based and ISS microgravity testing. Specifically, we address the RL Sim2Real gap for these robots operating on-board the ISS. 

Astrobee is a 12.5 inches (30.5 cm) cube-shaped robot designed to conduct intravehicular activities (IVAs) on-board the ISS. 
It uses electric fan propulsion for six degree-of-freedom (DOF) motion, and is equipped with cameras, sensors, and a 2-DOF perching arm for inspection and servicing tasks \cite{bualat2015astrobee}\cite{kang2024astrobee}. 
In addition to its operational capabilities it serves as a platform that enables researchers to test free-flyer algorithms and payloads rapidly in a zero-G pressurized environment \cite{Albee2020}. 
Recently, DRL using neural networks has emerged as a promising approach for enhancing spacecraft control, offering the potential for greater autonomy, adaptability, and optimized performance in challenging space environments \cite{elhariry2024drift}. 
Therefore we are applying RL for autonomous control of the Astrobee as the first step of an effort called Autonomous Planning In-space Assembly Reinforcement-learning free-flYer (APIARY) experiment \cite{2025_APIARY_iSpaRo_Paper_1} that is working towards autonomous robotic Assembly, Integration and Testing (AI\&T) for future In-Space Servicing, Assembly, and Manufacturing (ISAM) space missions. 
%
RL policies for robotics are typically trained and evaluated in simulation before being applied on hardware.
This poses the challenge of crossing the Sim2Real gap because simulations do not always perfectly capture the physical dynamics of the real world or unexpected variations for between the simulation and the real world hardware and environment \cite{Ranaweera_Mahmoud_2023}.
To overcome the Sim2Real gap we used the following workflow.
First, an RL policy was generated, through training within the high fidelity NVIDIA Omniverse Isaac Lab physics simulator, to output wrench commands to move the Astrobee in a zero-G environment to a desired pose.
NVIDIA Omniverse allows for training 10000s of parallel randomizable environments to maximize robotic experience during training. 
To more effectively train the RL policy against variations that could be introduced during testing, a curriculum \cite{narvekar2020curriculumlearningreinforcementlearning} was used to gradually introduce and train on such variations across the parallel environments.
Next, the RL policy was validated within the NASA Ames' Astrobee simulator using Gazebo and ROS Noetic, proving that the policy can replace the control of the robot's motion within the existing NASA Astrobee Robot Software. 
Next, preliminary real-world testing was conducted where the RL policy's wrench control commanded the Astrobee's fan-based propulsion system at the NASA Ames' Granite Lab, a terrestrial microgravity facility comprised of a large granite table that can mimic zero-G in a single plane by floating the Astrobee on air-bearings. 
The performance of the RL policy was compared to the baseline Astrobee controller. Finally, an RL policy was successfully tested in the microgravity environment of the ISS; this represents a first-of-its-kind demonstration of RL-based control for free-flying space robots. 
This successful implementation enables the development of complex, autonomous behaviors for Astrobee in the challenging microgravity environment of the ISS, unlocking new capabilities for scientific research and operational tasks. Furthermore, this novel methodology for training RL policies for space-based applications and verifying them terrestrially creates a pipeline for RL training to bridge the Sim2Real gap that will pave the way for future autonomous robot space missions.
These results highlight the transformative potential of RL for advancing robotic autonomy in space exploration and allowing for the possibility of rapid ground simulation RL training for on-orbit deployment to adapt to real-time mission needs.  

\section{Related Work} 
Spacecraft attitude control has traditionally relied on classical methods, such as Proportional-Derivative (PD) and Proportional-Integral-Derivative (PID) controllers, alongside bang-bang control for translational maneuvers~\cite{wertz2012spacecraft}.
Proportional output actuators—reaction wheels, rate-moment gyros, and magnetic torquer bars—are well-suited to these linear control strategies.
However the linear parameterization of dynamic uncertainty exploited by standard fixed-base adaptive controllers cannot be obtained when the base is free-floating \cite{sanner1995adaptivefreenn}.
Neural networks can be used to model the non-linear parameterization of the dynamic uncertainty for adaptive controllers and have demonstrated success for reliable adaptive control of space manipulators on a free floating base \cite{sanner1995adaptivefreenn}\cite{vance1996coordinatednn}\cite{wenhui2018freespacenn}.
The use of neural networks for adaptive control results in greater performance, and space fuel savings \cite{wenhui2018freespacenn}.

RL for spacecraft control has previously been explored in simulations for various tasks \cite{tipaldi2022Rl_space_control_review} \cite{charles2024ppo_docking_sim}. 
These tasks include autonomous lunar landing~\cite{scorsoglio2021meta_drl_autonomous_lunar_landing}, fuel efficient Mars landing~\cite{gaudet2014rl_fuel_efficient_mars_landing}, and spacecraft formation flying~\cite{dimauro2018gnc_autonomy_formation_flying}. 
DRL with Proximal Policy Optimization (PPO)~\cite{schulman2017ppo} has shown promising results for robotic control such as end-effector control of an arm connected to a free flying satellite~\cite{sah2024rl_rotation_control_floating}, and spacecraft debris collision avoidance~\cite{MU2024rl_spacecraft_collision_avoidance}. 
PPO has shown considerable success in bridging the sim-to-real gap, facilitating the deployment of policies trained in simulation onto real-world robotic platforms. Furthermore, PPO has been demonstrated in dexterous in-hand manipulation tasks~\cite{openai2019learningdexterousinhandmanipulation}, highlighting its robustness in sim-to-real transfer for complex manipulation tasks. 
This robustness extends to the control of floating platforms in simulated zero-g environments that were trained in a simulated environment and tested on the ground by floating robots on an epoxy floor~\cite{mahya2024ppo_control_zerog}. 
However, PPO is sample inefficient requiring extensive training samples with domain randomization for effective sim-to-real transfer. 
Nvidia's Omniverse GPU based physics simulator with IsaacLab~\cite{mittal2023orbit}, a framework for using reinforcement learning and imitation learning for robot training, alleviate the issues with massively parallelizable training environments and ability to extensively train with randomized domains across the environments with successful applications such as quadruped robot locomotion~\cite{rudin2022learning}, and furniture assembly~\cite{ankile2024assembly}. 
Inspired by the success of PPO in related robotic domains, we develop and evaluate a RL approach for Astrobee, aiming to advance the practical application of RL for spacecraft control and deployment in space.  

\section{Methods}

\subsection{Proximal Policy Optimization for Spacecraft Control}
\begin{figure}[h]
    \centering 
    \includegraphics[width=0.8\linewidth]{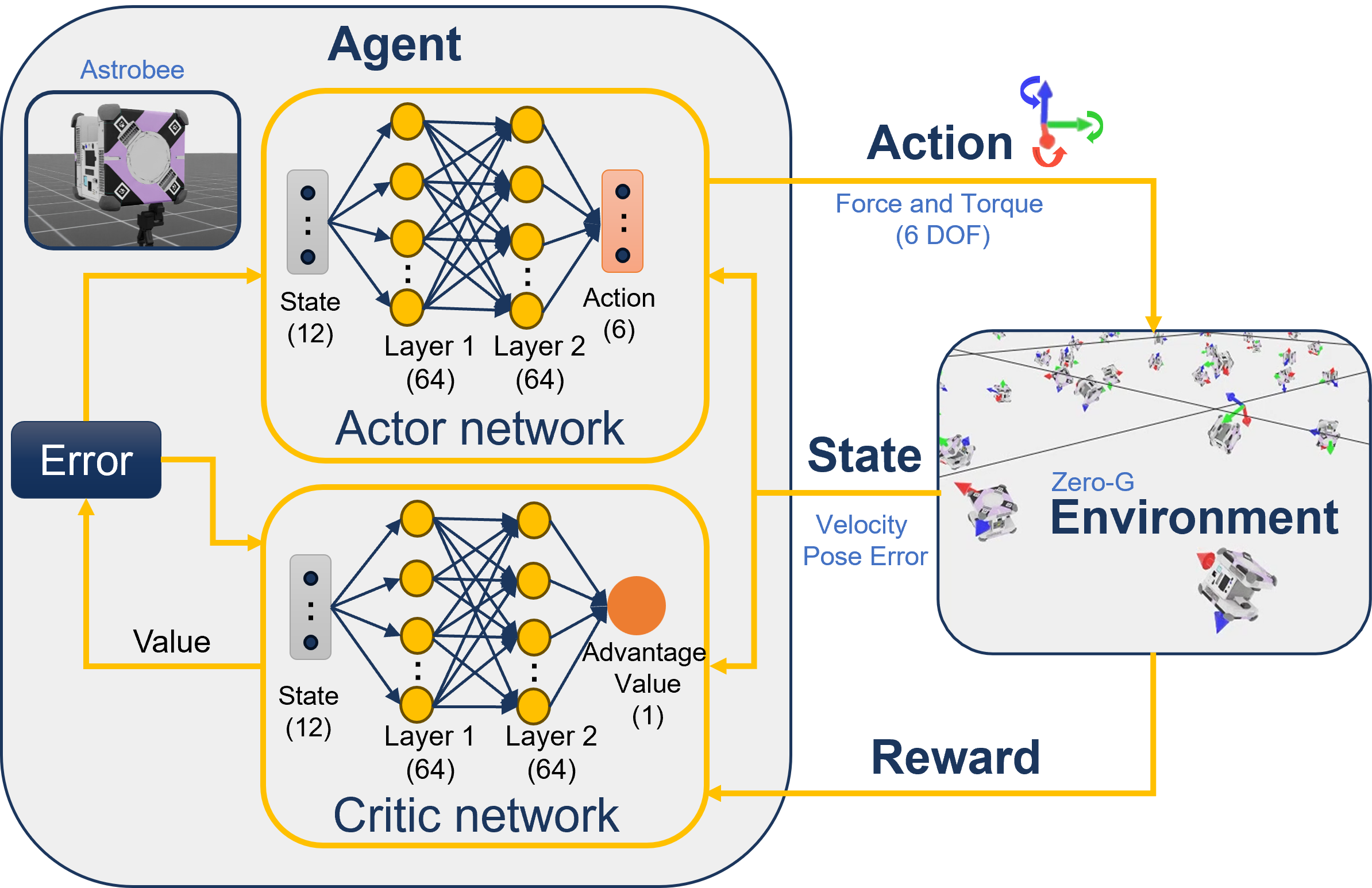}
    \caption{Astrobee Flight Control Reinforcement Learning with PPO}
    \label{fig:ppo} 
\end{figure}

PPO~\cite{schulman2017ppo} is a type of DRL actor-critic algorithm with two components: (1) an actor responsible for deciding which actions to take, and (2) a critic responsible for evaluating the actions taken by the actor.
PPO iteratively improves the policy through trial and error. 
The algorithm begins with an initial policy and then collects data by interacting with the environment and taking actions based on the current policy. 
For PPO training, the agent was the Astrobee robot in a simulated zero-G environment. The network's observations (inputs) consisted of the Astrobee's linear and angular velocities, along with its position and orientation errors relative to the goal pose.
The Astrobee's actions were to apply force and torque at its center of mass (COM) for movement.
The reward for the robot's actions are positive based on how close the Astrobee is to the goal pose and negative for large velocities. 
Figure \ref{fig:ppo} shows the RL training process used, and the actor and critic network architecture used.
The actor and critic are trained jointly with the same architecture and inputs but differ in their outputs with the actor outputting 6 values, 3 for the force and 3 for the torque, and the critic outputting a single value for the ``value'' of the action taken by the actor.
The network size is relatively small with each network having 2 hidden layers of 64 neurons which is necessary to fit on the Astrobee hardware which uses a Wandboard Dual board for low level control with our policy needing to take up less than 1MB of memory.

To train our PPO policies we use PPO with a clipped objective and Generalized Advantage Estimation (GAE)~\cite{schulman2018gae}, as follows:

\begin{equation}
\begin{aligned}
L^{\mathrm{CLIP}}(\theta)
= \hat{\mathbb{E}}_t \biggl[
\min\Bigl(
& \rho_t(\theta)\,\hat{A}_t,\;\\
& \operatorname{clip}\bigl(\rho_t(\theta),\,1-\epsilon,\,1+\epsilon\bigr)\,\hat{A}_t
\Bigr)
\biggr]
\end{aligned}
\end{equation}

where  $\rho_t(\theta) := \frac{\pi_\theta(a_t\mid s_t)}{\pi_{\theta_{\text{old}}}(a_t\mid s_t)}$ 
and $\pi_{\theta}(a_t|s_t)$ is the probability of taking action $a_t$ in state $s_t$ under the current policy with parameters $\theta$, with clipping at value $\epsilon$ and $\hat{A}_t$ is the advantage estimate computed using GAE as:

\begin{equation}
	\delta_t = r_t + \gamma V(s_{t+1}) - V(s_t)
\end{equation} 

\begin{equation}
	\hat{A}_t^{\text{GAE}(\gamma,\lambda)} = \sum_{l=0}^{\infty} (\gamma\lambda)^l \delta_{t+l}
\end{equation}

Where $\delta_t$ is the Temporal Difference (TD) residual, $V(s_t)$ is the estimated value function at state $s_t$, $r_t$ is the immediate reward at time step $t$, $\gamma$ is the discount factor for future rewards, and $\lambda$ is the GAE smoothing parameter. 
We use PPO with GAE in the NVIDIA Omniverse Isaac Lab physics simulator for its proven effectiveness in robotic control~\cite{rudin2022learning}~\cite{ankile2024assembly}, successful sim to real transfer capabilities~\cite{Muknahallipatna2024ppo_compare}, and efficient, rapidly testable training within the Isaac Lab environment. 
This approach was applied to motion optimization for reaching a desired pose, a fundamental functionality required to validate the RL policy on the Astrobee simulator and hardware on the granite table and on the ISS. 

\subsection{Training the PPO Policy}
A zero-G environment simulating the conditions on the ISS was created to train the PPO policy to perform Astrobee flight control. 
The training environment initializes the Astrobee in a suspended state, with PPO observations comprising of its linear and angular velocities and the position and orientation errors between its initial and goal poses all in 6 DOF. 
The trained policy generates commands for the robot's force and torque (wrench), mirroring the output of the Astrobee's Guidance Navigation and Control (GNC) controller~\cite{smith2015astrobee}.
The Astrobee GNC uses classical control to determine the thrust needed for the on-board fan-based propulsion system, with the goal of the PPO policy to replace the classical control. 
Rewards during training prioritize minimizing position and orientation error (positive rewards) and penalize excessive linear and angular velocity (negative rewards). 
Velocity penalties are calculated by summing the squared value of linear and angular velocities and multiplying them by a negative scaling factor. 
Position error is determined by subtracting the goal position from the current position. 
Orientation error, $q_{error}$, is calculated as the distance between the current, $q_{current}$, and goal, $q_{goal}$, quaternions as follows:

\begin{equation}
    q_{error} = \overline{q_{current}} \cdot q_{goal}
\end{equation}

where $\overline{q_{current}}$ is the conjugate of the current Astrobee orientation.

Both position and orientation error are then mapped onto the tanh function, as follows, to produce a positive reward that approaches one as the agent gets closer to the goal position and orientation:
\begin{equation}
	(1-tanh(error / goal_{\theta})) * scale
\end{equation}

Here the $goal_{\theta}$  is the threshold distance the Astrobee needs to be to start collecting the reward, which is the same as the Astrobee GNC threshold for determining if the Astrobee successfully moved to the desired pose, and $scale$ is a scaling factor for the reward.
The total reward is a weighted sum of these terms. 
The desired outcome of this reward function is that the robot will command the wrench action to move the robot toward the goal pose, minimizing position and orientation error, using only the minimally required linear and angular velocity for better controlled movement speed. 

Using the reward function, PPO, and network architecture shown in Figure \ref{fig:ppo}, three different reinforcement learning policies were trained.
The policies are generated from training 10,000 parallel environments and over 10,000 training iterations.
The first policy was trained with goal positions that are within a 1.0 meter radius sphere of the initial position with orientations given as a randomized quaternion with values in the range of (1.0, [0,0.5], [0,0.5], [0,0.5]) to simulate the range of typical maneuvers an Astrobee might make on the ISS and will be referred to as the \textit{no curriculum} policy. 
The other two policies were trained with curriculums which are described in the following section.

\subsection{Curriculum Training}

The curriculum method can be used during RL training to gradually increases difficulty or complexity of a task to improve learning efficiency, stability, and final performance~\cite{narvekar2020curriculumlearningreinforcementlearning}. 
A curriculum starts training the robot on an easy version of the desired task and then increases difficulty as the agent meets a desired milestone. 
Additionally, the environmental conditions can be varied across difficulty levels, such as varying the mass. 
Two RL policies were trained with a curriculum to gradually train more difficult pose error maneuvers and additionally add robustness to mass variation. 
The training was focused on mass variation because it was unknown whether during ISS testing whether the Astrobee would have a payload or not.
The RL policy needed to be able to handle either scenario and still complete the desired maneuvers. 

The curriculum training for the RL policies learning is divided into levels, with each level introducing more variation in the goal position, goal orientation, and Astrobee mass.
To advance to the next level an environment needs to maintain a reward greater than a set threshold for a certain number of time steps, in this case 550 and 1000 respectively with each time step lasting 0.016 seconds.
The initial level of the curriculum training had a constant end goal pose that is 0.5 meters away with the same orientation and using the normal mass of the Astrobee. 
The curriculum consisted of 22 levels total with each level increasing variation in the goal position, goal orientation, or the Astrobee mass.
Table \ref{tab:curriculum} shows the levels of the curriculum and the amount of variation at each level.  
Each level adds a variation of $V = random(-R,+R)$ where R is the floor or ceiling amount for the range of values to randomly add or subtract from the baseline value and V is the value to add to the baseline to give it variation. 
The baseline value for goal position is (x=0.5, y=0.5, x=0), the baseline for the goal orientation is q=(w=1.0, a=0, b=0, c=0), and the baseline value for mass $m$ is 9.0877kg. 
The (x, y, z) position values are all modified by $V$, for orientation only the (a,b,c) values of the quaternion are modified by $V$, and mass $m$ is modified by $V$. 
The randomization happens in each of the 10000 Astrobee training environments individually, allowing for a wide range of randomized values to be explored by the policy for more efficient learning and increased robustness.
Two policies were trained using the curriculum, one using the exact parameters of the curriculum specified that was tested in hardware on the ground and on the ISS, which will be called \textit{iss tested} and another that uses the same curriculum but has the Astrobee mass varied even more greatly from the beginning having values in the range $m = m + random(-9.0877, +9.0877)$ from level 1 onwards and demonstrated only in simulations to see if an RL policy is more robust to mass variation than classical control. 
This will me called \textit{more mass var}. 
 
\begin{table}[h]
\caption{Curriculum Training Levels}\label{tab:curriculum}
\centering
\begin{tabular}{|l|l|l|l|l|l|l|}
\hline
\textbf{Level}    & 1    & 2    & 3   & 4   & 5   & 6    \\ \hline
Position +/-  & 0    & 0    & 0   & 0.1 & 0.1 & 0.2  \\ \hline
Orientation +/- & 0    & 0.05 & 0.1 & 0.1 & 0.1 & 0.15 \\ \hline
Mass +/- & 0    & 0    & 0   & 0   & 0.5 & 0.5  \\ \hline \hline
\textbf{Level}    & 7    & 8    & 9   & 10  & 11  & 12   \\ \hline 
Position +/-  & 0.25 & 0.3  & 0.4 & 0.5 & 0.6 & 0.7  \\ \hline
Orientation +/- & 0.2  & 0.3  & 0.4 & 0.5 & 0.5 & 0.5  \\ \hline
Mass +/- & 0.5  & 0.5  & 0.5 & 0.5 & 0.5 & 0.5  \\ \hline \hline
\textbf{Level}    & 13   & 14   & 15  & 16  & 17  & 18   \\ \hline 
Position +/-  & 0.8  & 0.9  & 1.0 & 1.1 & 1.2 & 1.3  \\ \hline
Orientation +/- & 0.5  & 0.5  & 0.5 & 0.5 & 0.5 & 0.5  \\ \hline
Mass +/- & 0.5  & 0.5  & 0.5 & 0.5 & 0.5 & 0.5  \\ \hline \hline
\textbf{Level}    & 19   & 20   & 21  & 22  &     &      \\ \hline 
Position +/-  & 1.4  & 1.5  & 1.5 & 1.5 &     &      \\ \hline
Orientation +/- & 0.5  & 0.5  & 0.5 & 0.5 &     &      \\ \hline
Mass +/- & 0.5  & 0.5  & 1.0 & 1.5 &     &      \\ \hline
\end{tabular}
\end{table}


\section{Experiments and Results}
Experiments to validate the trained policies were done in 1) the same NVIDIA Omniverse simulation environment in which the policies were trained in, 2) a simulation environment of the ISS and Astrobee using NASA Ames software and classical controller, and 3) on the Astrobee hardware both on the ground and on the ISS in microgravity.
Only the \textit{iss tested} policy trained with the curriculum was able to be tested and validated in hardware.
The other two policies, the \textit{no curriculum} and \textit{more mass var} policies were only tested in simulations.
The  NASA Ames Research Center Astrobee was able to be experimentally tested on the ground at the NASA Ames Granite Lab and then in space on-board the ISS. 
The ground hardware was equivalent to the in-space Astrobee and allowed for initial validation of movements in 2 dimensions before flight testing.  

\subsection{Simulation}

Before hardware testing the policies were evaluated in the same NVIDIA Omniverse simulation environment the Astrobee RL control policies were trained in.
To evaluate the policies, 10000 environments and 10 seeds used for randomization were generated with randomized goal positions and orientations corresponding to $R=0.5$ for both position and orientation variance, the amount the \textit{no curriculum} policy is trained on.
Three separate evaluations corresponding to different mass variations of no variation,  $m =m + random(-2,+2)$, and $m =m + random(-9.0877,+9.0877)$ are done to evaluate the policies robustness to mass variation.
Each environment was rated a success if the policy was able to reach the goal position and goal orientation within the margin of error acceptable by the Astrobee GNC controller: 0.1m translation error with 20 degrees rotational error.
Results of the evaluation are shown in Table \ref{tab:omni}, with success rate shown as a percentage of the number of environments Astrobees that reached the goal position and orientation.
The results show that the \textit{no curriculum} policy had a very low success rate only reaching the goal position and orientation in $8.9\%$ to $9.77\%$ of environments. 
Interestingly the mass variance did not have a large impact on the success rate of the \textit{no curriculum} policy.
The curriculum trained policies were far more successful with consistently over $89\%$ success regardless of the mass variation demonstrating robustness to such variation.
The \textit{more mass var} policy performed slightly better than the \textit{iss tested} policy, likely because the policy was trained to handle more variation with higher mass variation throughout training.  
The RL policy shows consistent results across seeds with only small, less than 1\%, standard deviations.

\begin{table}[h]
\caption{Trained Policies Omniverse Simulation Results} \label{tab:omni}
\centering
\begin{tabular}{|l|l|l|}
\hline
Policy                               & Mass Variation (kg) & Success Rate \\ \hline
\multirow{3}{*}{\textit{no curriculum}}       & 0              & 9.77$\pm$0.43\%       \\ \cline{2-3} 
                                     & +/-2           & 9.74$\pm$0.32\%      \\ \cline{2-3} 
                                     & +/-9.0877      & 8.9$\pm$0.34\%      \\ \hline
\multirow{3}{*}{\textit{iss tested}}          & 0              & 91.99$\pm$0.17\%        \\ \cline{2-3} 
                                     & +/-2           & 91.89$\pm$0.14\%      \\ \cline{2-3} 
                                     & +/-9.0877      & 89.19$\pm$0.9\%      \\ \hline
\multirow{3}{*}{\textit{more mass var}} & 0              & 95.4$\pm$0.38\%       \\ \cline{2-3} 
                                     & +/-2           & 94.27$\pm$0.5\%      \\ \cline{2-3} 
                                     & +/-9.0877      & 92.08$\pm$0.67\%      \\ \hline
\end{tabular}
\end{table}

\begin{figure}[h]
    \centering 
    \includegraphics[width=0.8\linewidth]{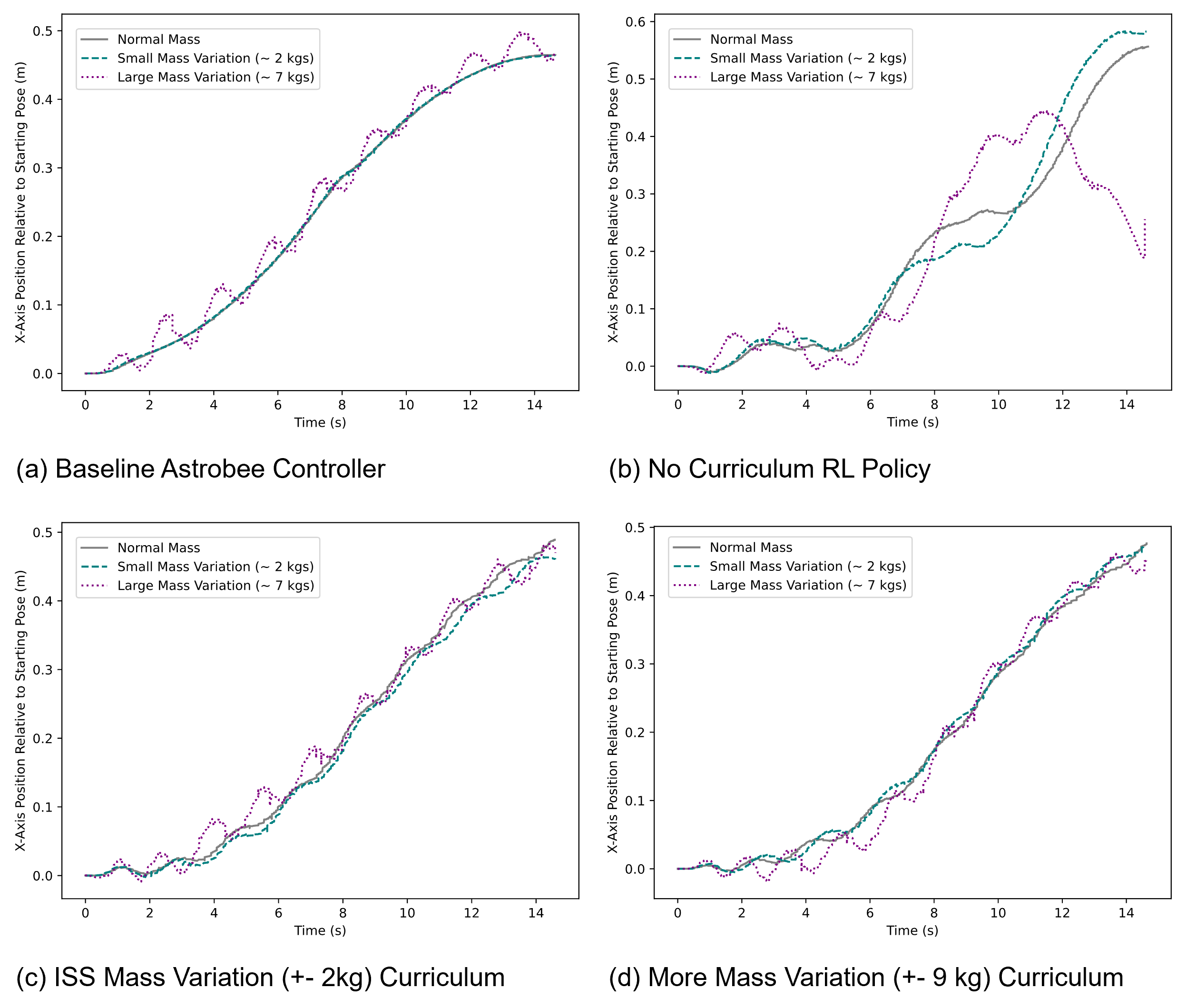}
    \caption{MSE comparison of baseline controller and various RL policies for robustness to mass variation: (a) Baseline Astrobee Controller, (b) \textit{no curriculum} RL Policy, (c) \textit{iss tested} Mass Variation (+- 1.5kg) Curriculum, (d) \textit{more mass var} (+- 9kg) Curriculum}
    \label{fig:sim_mass_variation_data} 
\end{figure}

The policies were also evaluated on the Gazebo simulation created by NASA Ames that uses the same code the Astrobee hardware uses for control and inputting commands.
Using the Gazebo version of the simulation, the policies can be directly compared to the Astrobee GNC controller, and since it is a simulation vary the mass of the Astrobee similar to the Omniverse simulation.
For a fair comparison the GNC control and policies were all given an undock maneuver to perform.
The Astrobee always begins docked in the Gazebo simulation guaranteeing the same start position and orientation in the simulation.
The undock maneuver is also used because it was performed and tested on hardware ground and ISS tests as well.
The undock maneuver is effectively a translation of 0.5 meters in the x direction.
Mass variation was also evaluated, with the Astrobee having the baseline mass, then a small mass variation of -2kg and a large mass variation of -7kg.
Figure \ref{fig:sim_mass_variation_data} shows the x position data of the Astrobee performing the undock, moving from the starting x position of 0 to 0.5 meters across the GNC controller and all policies with the mass variations tested.
Figure \ref{fig:sim_mass_variation_data} (a) shows the GNC controller undock, with a smooth curve towards the goal for the normal baseline mass and the small mass variation, but an oscillatory movement with the large mass variation. Figure \ref{fig:ground_mass_variation_data} (b) shows the \textit{no curriculum} policy either overshot or undershot the goal position, while the \textit{iss tested} policy in (c) and the \textit{more mass var} policy in (d) reached the goal position with less smooth curves than the GNC controller but similar performance on the large mass variation to the GNC controller.
To better measure the robustness of the policies to mass variation given the x position data, Table \ref{tab:mass_var_table} shows the Mean Squared Error (MSE) between the small and large mass variation curves compared to the normal baseline mass for the undock, with the MSE indicating the distance of the variation curves from the baseline with a value of 0 being without error and larger values showing greater distance, or error, from the baseline for each controller where the normal mass is used.
While the GNC controller is fairly robust to mass variation, the curriculum trained policies MSE values show they had similar robustness to mass variations giving more confidence to testing on the ISS in microgravity.

\begin{table}[h]
\caption{} \label{tab:mass_var_table}
\resizebox{\columnwidth}{!}{%
\begin{tabular}{|l|c|ccc|}
\hline
\multicolumn{1}{|c|}{} &  & \multicolumn{3}{c|}{RL Policy Controller} \\ \cline{3-5} 
\multicolumn{1}{|c|}{\multirow{-2}{*}{\begin{tabular}[c]{@{}c@{}}Mass Variation \\ of Test\end{tabular}}} & \multirow{-2}{*}{\begin{tabular}[c]{@{}c@{}}Baseline \\ Controller\end{tabular}} & \multicolumn{1}{c|}{\begin{tabular}[c]{@{}c@{}}No \\ Curriculum\end{tabular}} & \multicolumn{1}{c|}{\begin{tabular}[c]{@{}c@{}}ISS Mass Variation \\ (+- 2kg) Curriculum\end{tabular}} & \begin{tabular}[c]{@{}c@{}}More Mass Variation \\ (+- 9 kg) Curriculum\end{tabular} \\ \hline
{\color[HTML]{029386} \begin{tabular}[c]{@{}l@{}}Error: Small Mass \\ Variation ($\sim$2 kgs)\end{tabular}} & \cellcolor[HTML]{63BE7B}0.00 & \multicolumn{1}{c|}{\cellcolor[HTML]{FFE283}0.11} & \multicolumn{1}{c|}{\cellcolor[HTML]{FFEB84}0.04\%} & \cellcolor[HTML]{7FC67C}0.01\% \\ \hline
{\color[HTML]{7E1E9C} \begin{tabular}[c]{@{}l@{}}Error: Large Mass \\ Variation ($\sim$7 kgs)\end{tabular}} & \cellcolor[HTML]{E3E382}0.03\% & \multicolumn{1}{c|}{\cellcolor[HTML]{F8696B}1.10\%} & \multicolumn{1}{c|}{\cellcolor[HTML]{F8E983}0.04} & \cellcolor[HTML]{FFEB84}0.04 \\ \hline
\end{tabular}%
}
\end{table}

\subsection{Ground Testing: Mass Variation}
\begin{figure}[h]
    \centering 
    \includegraphics[width=0.8\linewidth]{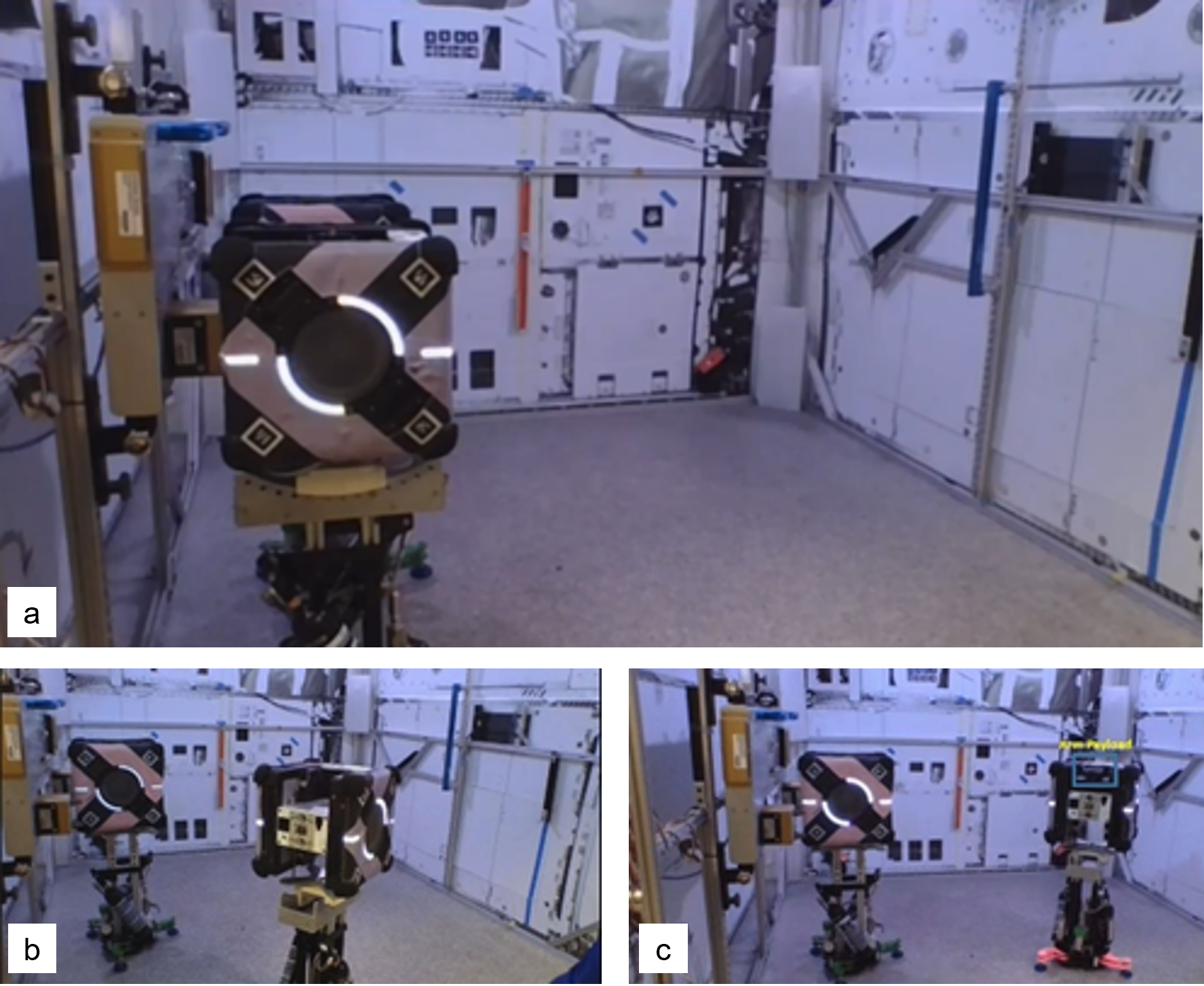}
    \caption{Ground testing of RL control on Astrobee hardware at NASA Ames Granite Lab. (a) Docked state with robot without a payload. (b) Undocked state of robot without a payload (c) Robot including the robotic arm payload attached.}
    \label{fig:ground_mass_variation_images} 
\end{figure}

The \textit{iss tested} policy was tested on the Astrobee hardware at the NASA Ames Granite Lab.
Testing at the NASA Ames Granite Lab allowed for 3 degree-of-freedom (DOF) validation of the Astrobee RL control. Initial validation to assure that the RL control could complete the desired maneuvers within the goal pose tolerances were conducted on the ground before ISS testing. 
Additionally, testing on the ground allowed for experimental validation of mass variation robustness of the RL policy due to the curriculum training. 
Figure \ref{fig:ground_mass_variation_images} shows the Astrobee with and without the arm payload.  

\begin{figure}[h]
    \centering 
    \includegraphics[width=0.8\linewidth]{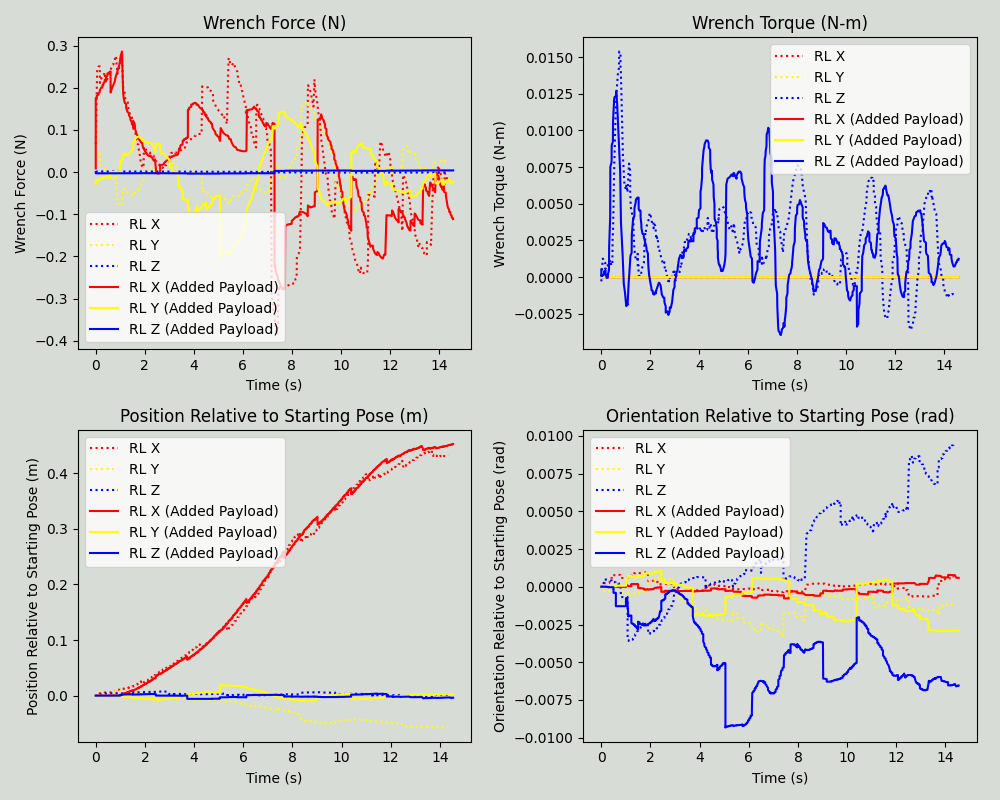}
    \caption{Dataset comparing undock with (solid lines) and without (dotted lines) the arm mass payload.}
    \label{fig:ground_mass_variation_data} 
\end{figure}

The Astrobee was able to perform successful maneuvers with and without the arm payload using the \textit{iss tested} RL policy. 
Figure \ref{fig:ground_mass_variation_data} shows the undocking maneuvers being successfully executed with and without the arm mass payload. 
This validates that the curriculum was able to successful train an RL policy robust to mass variation in hardware. 

\subsection{Flight Testing}

\begin{figure}[h]
    \centering 
    \includegraphics[width=0.8\linewidth]{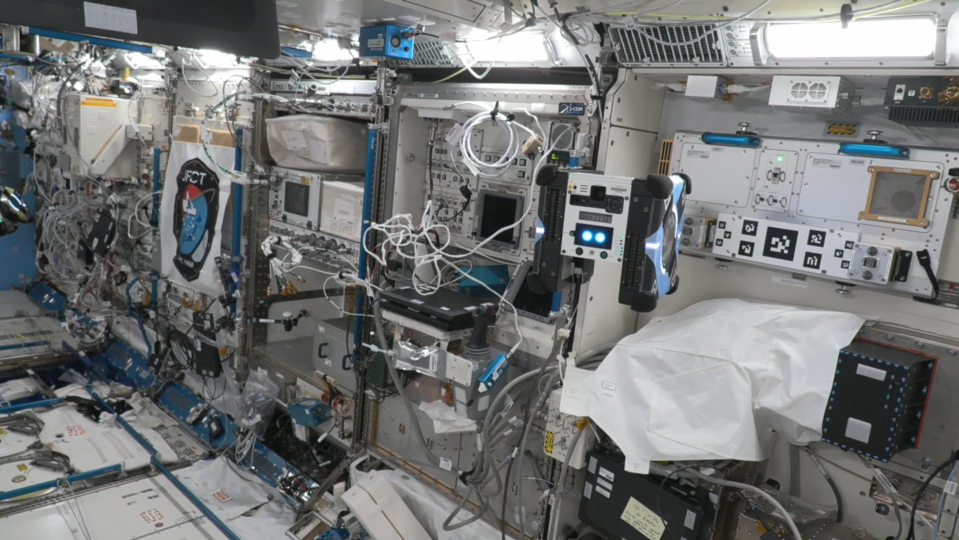}
    \caption{Astrobee free-flyer under RL control in zero-G on-board the ISS.}
    \label{fig:flight_testing} 
\end{figure}

\begin{figure}[h]
    \centering
    \includegraphics[width=0.8\linewidth]{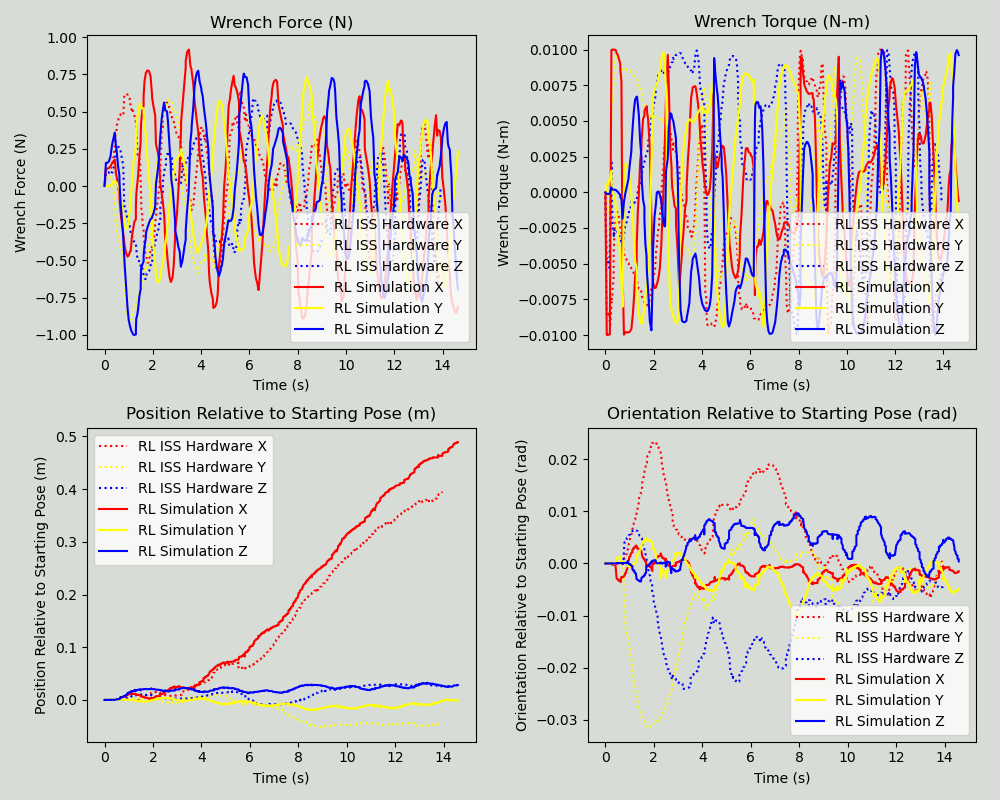}
    \caption{Comparison of undock performance in the ISS simulation vs the actual ISS with Astrobee hardware running RL control.}
    \label{fig:flight_vs_sim} 
\end{figure}

Following simulation and ground testing validation the curriculum trained \textit{iss tested} RL policy was deployed on the NASA Astrobee in a microgravity enviornment on-board the ISS, shown in Figure \ref{fig:flight_testing}. 
The RL policy controlled the 6 DOF pose of the Astrobee's free-flying motion to complete maneuvers such as undock, rotate, translation, and redocking \cite{2025_APIARY_iSpaRo_Paper_1}. 
Figure \ref{fig:flight_vs_sim} shows the comparison of the undock maneuver of the real hardware flight test on the ISS versus the Gazebo simulation undock with both using the \textit{iss tested} RL control policy.
While the hardware undock did undershoot the 0.5m target, it was still within the acceptable margin of error for a successful undock.
The success of this first RL autonomous control of a free flyer in microgravity demonstrates the successful crossing of the Sim2Real gap between training in simulation to testing in hardware on the target hardware and environment paving the way for future autonomous spacecraft control.

\subsection{Conclusion}

This Astrobee ground and ISS testing in hardware successfully demonstrated RL based autonomous control of an in-space free-flyer for the first time. 
Training with a curriculum was successfully shown to be robust to mass variation during ground hardware testing on a Granite table. 
This experiment has demonstrated a pipeline from sim to hardware for RL control of space robots.
Therefore we believe our other simulation results would have been successfully validated on the hardware on the ISS had there been time to test them there.

Future work can build upon this initial validation of the AI\&T process to cross the Sim2Real gap between simulation and ground testing to space deployment of autonomous robots. 
Using the training and testing methods developed here, more complex policies can be generated for required in-space tasks. 

\section{Acknowledgments}
Thanks to ONR for supporting our research. Thanks to the NASA Ames Research Center Astrobee team for helping us perform the Granite Lab testing: Ruben Garcia Ruiz, Jordan Kam, Katie, Hamilton, and Roberto Carlino and thanks to the rest of the NASA Astrobee team for aid coordinating in and allowing us to do ISS testing: Jonathan Barlow, Henry Orosco, Andres Mora Vargas, Jose Benavides, Aric James Katterhagen, and Simeon Kanis. Special thanks to Kirk Hovell and the rest of the Obruta Space Solutions team for allowing us to be part of their AstroSee ISS test.

\bibliographystyle{IEEEtran}
\bibliography{example} 

\end{document}